\documentclass[runningheads]{llncs}
\usepackage{graphicx}
\usepackage{url}            
\usepackage{booktabs}       
\usepackage{amsfonts}       
\usepackage{nicefrac}       
\usepackage{microtype}      
\usepackage{tikz}
\usepackage{tabularx}
\usepackage{hhline}
\usepackage{adjustbox}
\usepackage{multirow}
\usepackage{mathtools}
\usepackage{bbm}
\usepackage{color, colortbl}

\usepackage{cellspace}
\DeclarePairedDelimiterX\set[1]\{\}{\nonscript\,#1\nonscript\,}

\usetikzlibrary{shapes.geometric, arrows}
\tikzstyle{startstop} = [rectangle, rounded corners, minimum width=1.5cm, minimum height=0.8cm,text centered, draw=black, fill=red!30]
\tikzstyle{io} = [trapezium, trapezium left angle=70, trapezium right angle=110, minimum width=1.5cm, minimum height=0.8cm, text centered, draw=black, fill=blue!30]
\tikzstyle{process} = [rectangle, minimum width=1.5cm, minimum height=0.8cm, text centered, draw=black, fill=orange!30]
\tikzstyle{decision} = [diamond, minimum width=1.5cm, minimum height=0.8cm, text centered, draw=black, fill=green!30]
\tikzstyle{block} = [rectangle, rounded corners, minimum width=1.5cm, minimum height=0.8cm,text centered, draw=black, fill=orange!30]
\tikzstyle{network} = [rectangle, rounded corners, minimum width=1.5cm, minimum height=0.8cm,text centered, draw=black, fill=green!30]
\tikzstyle{bim} = [rectangle, rounded corners, minimum width=0.1cm, minimum height=0.1cm, text centered, draw=blue, dash pattern=on 4pt off 4pt]
\tikzstyle{arrow} = [thick,->,>=stealth]

\usepackage{makecell}

\newcommand\tf[1]{\textbf{#1}}

\def\ie{\textit{i.e.}}
\def\eg{\textit{e.g.}}
\def\etal{{\textit{et~al.}}}

\newcommand{\myparagraph}[1]{\vspace{1pt}\noindent{\bf{#1}}~~}

\usepackage{colortbl}
\definecolor{lightgray}{gray}{0.75}
\definecolor{lightergray}{gray}{0.85}
\definecolor{Blue}{RGB}{3, 31, 97}
\definecolor{Blue1}{RGB}{214, 235, 245}
\definecolor{Blue2}{RGB}{235, 245, 250}
\definecolor{Gray}{RGB}{247, 252, 255}

\definecolor{convcolor}{HTML}{412F8A}
\definecolor{resnetcolor}{HTML}{8DA0CB}
\definecolor{vitcolor}{HTML}{fc8e62}

\begin{document}
\title{Momentum Contrastive Voxel-wise Representation Learning for Semi-supervised Volumetric Medical Image Segmentation}
\titlerunning{Momentum Contrastive Voxel-wise Representation Learning}
%
\author{Chenyu You\inst{1}
\and Ruihan Zhao\inst{2}
\and Lawrence Staib\inst{1}
\and James S. Duncan\inst{1}}

\authorrunning{C. You et al.}

\institute{Yale University, New Haven CT 06520, USA\\ 
\and
The University of Texas at Austin, Austin TX 78712, USA}

\maketitle              

\begin{abstract}
Contrastive learning (CL) aims to learn useful representation without relying on expert annotations in the context of medical image segmentation. Existing approaches mainly contrast a single positive vector (\ie, an augmentation of the same image) against a set of negatives within the entire remainder of the batch by simply mapping all input features into the same constant vector. Despite the impressive empirical performance, those methods have the following shortcomings: (1) it remains a formidable challenge to prevent the collapsing problems to trivial solutions; and (2) we argue that not all voxels within the same image are equally positive since there exist the dissimilar anatomical structures with the same image. In this work, we present a novel \textbf{C}ontrastive \textbf{V}oxel-wise \textbf{R}epresentation \textbf{L}earning (CVRL) method to effectively learn low-level and high-level features by capturing 3D spatial context and rich anatomical information along both the feature and the batch dimensions. Specifically, we first introduce a novel CL strategy to ensure feature diversity promotion among the 3D representation dimensions. We train the framework through bi-level contrastive optimization (\ie, low-level and high-level) on 3D images. Experiments on two benchmark datasets and different labeled settings demonstrate the superiority of our proposed framework. More importantly, we also prove that our method inherits the benefit of hardness-aware property from the standard CL approaches. Codes will be available soon.


\end{abstract}

    \begin{figure}[t!]
    	\centering 
        \includegraphics[width=\linewidth]{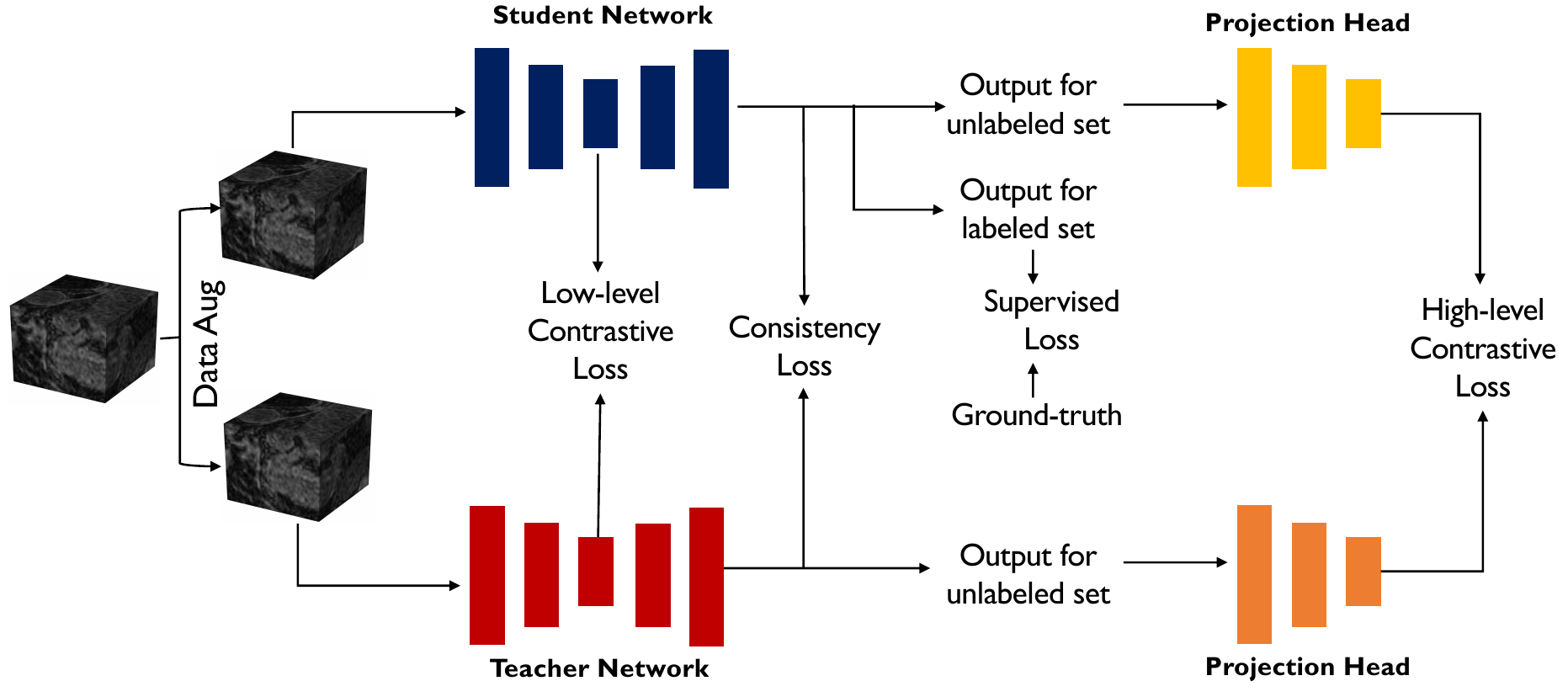}
        \vspace{-20pt}
        \caption{Overview of CVRL architecture. We learn rich dense voxel-wise representations by exploiting high-level context between voxels and volumetric regions and low-level correlations in both batch and feature dimensions in a semi-supervised manner.} 
        \label{fig:model}
        \vspace{-15pt}
    \end{figure}

\section{Introduction}

Learning from just a few labeled examples while leveraging a large amount of unlabeled data is a long-standing pursuit in the machine learning community, which is especially crucial for the medical imaging domain. Generating reliable manual annotations of 3D imaging data at scale is expensive, time-consuming, and may require domain-specific expertise. Due to privacy concerns, another challenge in medical imaging is relatively small training datasets.

In the medical imaging domain, substantial efforts~\cite{zhang2017deep,li2018semi,nie2018asdnet,yu2019uncertainty,bortsova2019semi,luo2020semi,chen2019multi,bai2019self} have been devoted to incorporating unlabeled data to improve network performance due to the limited 3D data and annotations. The most common training techniques are adversarial learning and consistency loss as regularization terms to encourage unsupervised mapping. Recently, contrastive learning (CL) has drawn considerable attention to learning useful representations without expert supervision and shown remarkable performance in the medical image analysis domain \cite{chaitanya2020contrastive,hu2021semi,you2021simcvd}. The central idea \cite{hadsell2006dimensionality,wu2018unsupervised,misra2020self,chen2020simple,tian2019contrastive,chaitanya2020contrastive} is to learn powerful representations invariant to data augmentations that maximize the agreement between instance embeddings from different augmentations of the same images. The major stream of subsequent work focuses on the choice of dissimilar pairs, which determine the quality of learned representations. The loss function used to contrast is chosen from several options, such as InfoNCE \cite{oord2018representation}, Triplet~\cite{wang2015unsupervised}, and so on. However, while remarkable, those methods assume that the repulsive effect of negatives can avoid collapsing along all dimensions to trivial solutions by explicitly using positive and negative pairs in the loss function. However, it has been empirically observed that such design may still collapse along certain dimensions (\ie, \textit{dimensional collapse}), as noted in \cite{hua2021feature,wang2021understanding}. Such scenarios can happen in predefined augmentations, which usually lead to better performance due to the inter-instance constraints while usually ignoring anatomically feasibility in the transformation.

\begin{figure}[t]
\centering
\includegraphics[width=0.8\linewidth]{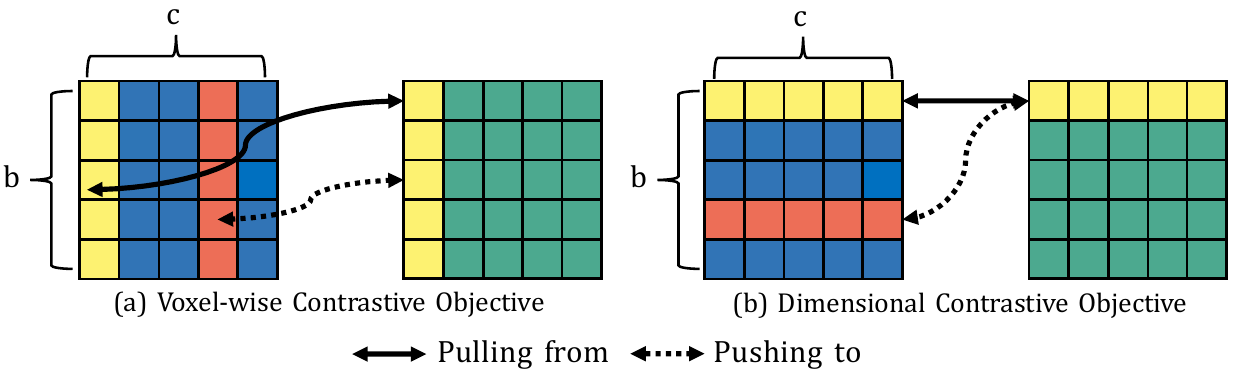}
\vspace{-5pt}
\caption{Comparison of (a) Voxel-wise Contrastive Objective (\ie, batch dimension), and (b) Dimensional Contrastive Objective (\ie, feature dimension). $b$ denotes batch size and $c$ is the feature dimension.} 
\label{fig:contrast}
\vspace{-10pt}
\end{figure}

In this paper, we present CVRL, a novel end-to-end semi-supervised framework to learn high-level contexts and local-level features in both the batch and the feature directions for 3D medical image segmentation. One \tf{blessing} comes from the recent finding \cite{jing2021understanding} in the context of image classification. The authors note that applying strong augmentation along the feature dimension may result in the dimensional collapse in CL. In other words, the augmented images are not ``standardized" well and may easily admits collapsed solutions (\eg, generating the same vector for all 3D scans), making it challenging or even infeasible in real-world clinical practice. Three \tf{key aspects} distinguish from the recent success \cite{chaitanya2020contrastive}. First, the standard CL encourages instance-level representation diversity within the batch. By contrast, we propose an anatomy-informed regularization among the feature dimension, as an intra-instance constraint to encourage feature diversity for its improved robustness, as illustrated in Figure \ref{fig:contrast}. This design is appealing: (1) our idea is rather plug-and-play and can be easily compatible with existing inter-instance constraints; and (2) it inherits the strength of CL in learning useful representations to improve the feature spaces’ discriminative capability (See Appendix). Second, we propose to perform low-level contrast in a lower-dimensional 3D subspace, which could capture rich anatomic information; and (3) existing methods mainly perform local contrast in the image-level space, which may usually lead to sub-optimal segmentation quality due to the lack of spatial information. In contrast, if the proposed method can learn more generic representation from 3D context, it will unlock the appealing prospect of 3D nature in medical images (\ie, 3D volumetric scans). We propose an additional high-level contrast to exploit distinctive features in the embedding feature space by designing a new 3D projection head to encode 3D features. We also theoretically show our dimensional contrastive learning inherits the hardness-aware property in Appendix. The results demonstrate that our segmentation network outperforms the state-of-the-art methods on two benchmark datasets, and generates object segmentation with high-quality global shapes.

\subsection{Overview}\label{subsec:overview}

An overview of the architecture is illustrated in Figure~\ref{fig:model}. Our CVRL is based on GCL \cite{chaitanya2020contrastive}, and follows its most important components such as data augmentations. our goal is to learn stronger visual representations that avoid collapsing for improving the overall segmentation quality with limited annotation clinical scenarios. In the limited annotation setting, we train semi-supervised CVRL alongside two components - supervised and unsupervised learning objectives. Specifically, we propose a novel voxel-wise representation learning algorithm to learn low-level and high-level representations from 3D unlabeled data by regularizing the embedding space and exploring the geometric and spatial context of training voxels. 

In our problem setting, we consider a set of training data (3D images) including $N$ labeled data and $M$ unlabeled data, where $N\ll M$. For simplicity of exposition, we denote limited label data as $\mathcal{D}_{l}=\left\{\left(\mathbf{x}_{i}, y_{i}\right)\right\}_{i=1}^{N}$, and abundant unlabeled data as $\mathcal{D}_{u}=\left\{\left(\mathbf{x}_{i}\right)\right\}_{i=N+1}^{N+M}$, where~$\mathbf{x}_{i} \in \mathbb{R} ^{H \times W \times D}$ are volume inputs, and $\mathbf{y}_{i} \in \{0, 1\}^{H \times W \times D}$ are ground-truth labels. Specifically, we adopt V-Net~\cite{yu2019uncertainty} as the network backbone~$F(\cdot)$, which consists of an encoder network and a decoder network. To maximize mutual information between latent representations, we design a projection head, that comprises one encoder network which share the similar architecture to the prior encoder network.

\subsection{Unsupervised Contrastive Learning}\label{sec:contrat}
A key component of CVRL is the ability to capture rich voxel-wise representations of high dimensional data by contrastive distillation. CVRL trains on the contrastive objective as an auxiliary loss during the volume batch updates. We utilize two contrastive learning objectives: (i) voxel-wise contrastive objective (ii) dimensional contrastive objective, each applied on a low-level feature and a high-level feature. The resulting combined training objective greatly improves the quality of learned representations.

First we establish some notations which will assist in explaining our approach. We denote a batch of input image $x_1 \cdots x_b$, the teacher encoder network $f$, the student encoder network $g$ and a set of data augmentation transformations $\mathcal{T}$ (\eg \, random flipping, random rotation, random brightness, random contrast, random zooming, cube rearrangement and cube rotation). Here $f$ and $g$ use the same encoder architecture $e(\cdot)$ as introduced in section \ref{subsec:overview}, but with different parameters. $z^f \in \mathbb{R}^{h \times w \times d \times c}$ = $f(t(x)))$ is a feature volume produced by the student encoder, whereas $z^g = g(t'(x))$ is the corresponding feature volume produced by the teacher encoder, under a different random transformation $t' \neq t$.

\myparagraph{Voxel-wise Contrastive Objective}
Using standard contrastive learning, we encourage the feature extractor to produce representations that are invariant under data augmentations. On the other hand, features should still preserve locality: different voxels in a feature volume should contain their unique information. Specifically, as the learned feature volumes are divided into slices, we pull the pairs of voxels that come from two augmentations of the same image closer; voxels at different locations or from different images are pushed away. To learn a feature extractor that unlocks the desired properties, we use InfoNCE loss~\cite{oord2018representation}:
\begin{equation}\label{eq:NCE}
\mathcal{L}_q = -\log\frac{\exp(q \cdot k_+/\tau)} {\exp(q \cdot k_+/\tau) + \sum_{k \in \mathcal{K}_-}\exp(q \cdot k/\tau)}
\end{equation}
The query $q \in \mathbb{R}^c$ is a voxel in a student feature volume $z^f$, and key $k \in \mathbb{R}^c$ comes from the teacher feature volume $z^g$. In particular, $k_+$, the positive key, is the teacher feature voxel corresponding to the same location in the same image as the query $q$. The set $\mathcal{K}_-$ contains all other keys in the mini-batch, from different locations and different inputs. $\tau$ is a temperature hyper-parameter.

To obtain the voxel-wise contrastive loss, we take the average over the set of query voxels $\mathcal{Q}_v$ consisting of all feature voxels in the mini-batch of student feature volumes:
\begin{equation}\label{eq:vco}
\mathcal{L}_v = \frac{1}{|\mathcal{Q}_v|} \sum_{q \in \mathcal{Q}_v}\mathcal{L}_q
\end{equation}

\subsubsection{Dimensional Contrastive Objective}
Motivated by recent findings on dimensional collapse in contrastive learning \cite{jing2021understanding,hua2021feature}, we propose a dimensional contrastive objective to encourage different dimensions/channels in the feature voxels to contain diverse information. Given a batch of student feature volumes of shape $b\!\times\!h\!\times\!w\!\times\!d\!\times\!c$, we group the first 4 dimensions to obtain a set of dimensional queries: $q \in \mathcal{Q}_d \subset \mathbb{R}^{(b \times h \times w \times d)}$ where $|\mathcal{Q}_d| = c$, the number of channels in the feature volume. We define the $\mathcal{K} = \{k_+\} \cup \mathcal{K_-}$ in the same way, but using the corresponding batch feature volumes from the teacher encoder. In the dimensional contrastive setting, $k_+$ is defined as the key vector that corresponds to the same feature dimension as the query $q$. The dimensional contrastive loss is the average over all query dimensions:
\begin{equation}\label{eq:dco}
\mathcal{L}_d = \frac{1}{c} \sum_{q \in \mathcal{Q}_d}\mathcal{L}_q
\end{equation}
We theoretically show our dimensional contrastive learning inherits the hardness-aware property in Appendix.

\myparagraph{Consistency Loss $\mathcal{L}_{c}$}
Recent work~\cite{laine2016temporal,tarvainen2017mean} show that using an exponential moving average (EMA) over network parameters is empirically shown to improve training stability and models' final performance. With this insight, we introduce an EMA teacher model with parameters~$\theta$ as the moving-average of the parameters~${\theta'}$ from the original student network. Specifically, the architecture of EMA model follows the original model. At training step $t$, the update rule follows $\theta_t = m\theta_{t-1} + (1-m)\theta'_t$, where $m\in [0,1)$ is momentum parameter. On the unlabeled set, we perform different perturbation operations on the unlabeled input volume sample~$x^{u}$,~\eg~adding noise~$\epsilon$. To encourage training stability and performance improvements, we define consistency loss as:
\begin{equation}\label{eq:mean-tea}
    \mathcal{L}^{\text{con}}=\mathcal{L}_{\mathrm{mse}}\left(g\left(x^{u};\theta,\epsilon\right), f\left(x^{u};\theta^{\prime},\epsilon^{\prime}\right)\right),
    \vspace{-3pt}
\end{equation}
where $\mathcal{L}_{\mathrm{mse}}$ is the mean squared error loss.

\myparagraph{Overall Training Objective} 
Our overall learning objective is to minimize a combination of supervised and unsupervised losses. On the labeled data, the supervised training objective is the linear combination of cross-entropy loss and dice loss. On the unlabeled dataset, unsupervised training objective consist of the consistency loss $\mathcal{L}_{c}$, high-level contrastive loss $\mathcal{L}^{\text{high}}$ (\ie, the linear combination of $\mathcal{L}_{v}^{\text{high}}$ and $\mathcal{L}_{d}^{\text{high}}$), and low-level contrastive loss $\mathcal{L}^{\text{low}}$ (\ie, the linear combination of $\mathcal{L}_{v}^{\text{low}}$ and $\mathcal{L}_{d}^{\text{low}}$). The overall loss function is:
\begin{equation}\label{eq:overall_obj}
    \mathcal{L} = \mathcal{L}^{\mathrm{sup}} + \alpha \mathcal{L}^{\text{high}} +  \beta \mathcal{L}^{\text{low}} + \gamma \mathcal{L}^{\text{con}}
    \vspace{-3pt}
\end{equation}
where $\alpha$, $\beta$, $\gamma$ are hyperparameters that balance each term.

\section{Experiments}\label{sec:exp}     
    \begin{table}[t]
        \caption{Quantitative segmentation results on the LA dataset. The backbone network of all evaluated methods are V-Net.}
        \label{tab:la_16}
        \centering
        \resizebox{0.95\textwidth}{!}{
        \begin{tabular}{c|c|c|c|c|c|c} \hline \hline
        \multirow{2}{*}{Method} & \multicolumn{2}{c|}{\# \textbf{scans used}} & \multicolumn{4}{c}{\textbf{Metrics}}  \\ \cline{2-7}
                                & Labeled        & Unlabeled        & Dice{[}\%{]} & Jaccard{[}\%{]} & ASD{[}voxel{]} & 95HD{[}voxel{]} \\ \hline
        V-Net \cite{milletari2016v}       
        & 80             
        & 0                
        & 91.14        
        & 83.82           
        & 1.52           
        & 5.75            
        \\ \hline
        V-Net              
        & 16             
        & 0                
        & 86.03        
        & 76.06           
        & 3.51           
        & 14.26    
        \\ \hline  \hline
        DAN \cite{zhang2017deep}                     & 16             & 64               & 87.52        & 78.29           & 2.42           & 9.01            \\ \hline
        DAP \cite{zheng2019semi}    
        & 16    
        & 64    
        & 87.89 	
        & 78.72  
        & 2.74   
        & 9.29  
        \\ \hline
        UA-MT \cite{yu2019uncertainty}              
        & 16             
        & 64               
        & 88.88        
        & 80.21           
        & 2.26           
        & 7.32            
        \\ \hline
        LG-ER-MT~\cite{hang2020local}         
        & 16             
        & 64               
        & 89.56        
        & 81.22           
        & 2.06           
        & 7.29            
        \\ \hline
        SASSNet~\cite{li2020shape}          
        & 16             
        & 64               
        & 89.27        
        & 80.82           
        & 3.13           
        & 8.83            
        \\ \hline
        Chaitanya \etal \cite{chaitanya2020contrastive} 
        & 16             
        & 64               
        & {89.94}        
        & {81.82}           
        & {2.66}           
        & {7.23}            
        \\ \hline
        CVRL(ours)          
        & 16             
        & 64               
        & \textbf{90.45}        
        & \textbf{83.02}        
        & \textbf{1.81}        
        & \textbf{6.56}           
        \\ \hline\hline
        V-Net \cite{milletari2016v}                   & 8                                           & 0                                    
        & 79.99          
        & 68.12           
        & 5.48           
        & 21.11           
        \\ \hline
        DAN \cite{zhang2017deep}                     
        & 8             
        & 72               
        & 80.87        
        & 70.65           
        & 3.72           
        & 15.96            \\ \hline
        DAP \cite{zheng2019semi}    
        & 8    
        & 72    
        & 81.89 	
        & 71.23 	
        & 3.80 	
        & 15.81         
        \\ \hline
        UA-MT \cite{yu2019uncertainty} 
        & 8                                           & 72                                   
        & 84.25          
        & 73.48           
        & 3.36           
        & 13.84           
        \\ \hline
        LG-ER-MT~\cite{hang2020local}               
        & 8                                           & 72                                   
        & 85.43          
        & 74.95           
        & 3.75           
        & 15.01           
        \\ \hline
        SASSNet~\cite{li2020shape}          
        & 8                                           & 72                                   
        & 86.81          
        & 76.92           
        & 3.94           
        & 12.54           
        \\ \hline
        Chaitanya \etal \cite{chaitanya2020contrastive}          
        & 8             
        & 72               
        & {84.95}        
        & {74.77}           
        & {3.70}           
        & {10.68}            
        \\ \hline
        CVRL(ours)          
        & 8                                           & 72                                   
        & \textbf{88.56}         
        & \textbf{78.89}           
        & \textbf{2.81}         
        & \textbf{8.22}         
        \\ \hline\hline
        \end{tabular}}
	\vspace{-10pt}
    \end{table}

\myparagraph{Dataset and Pre-processing}
We conduct our experiments on two benchmark datasets: the Left Atrium (LA) dataset from Atrial Segmentation Challenge\footnote{http://atriaseg2018.cardiacatlas.org/} and the NIH pancreas CT dataset \cite{roth2016data}. The LA dataset includes 100 3D gadolinium-enhanced MR imaging scans with annotations. The isotropic resolution of the scan is of $0.625\times 0.625 \times 0.625 \text{mm}^3$. We use 80 scans for training, and 20 scans for evaluation. For pre-processing, we crop all the scans at the heart region and normalized them into zero and unit variance with the size of sub-volumes $112\times 112\times 80 \text{mm}^3$. The pancreas dataset consists of 82 contrast-enhanced abdominal CT scans. We use the same experimental setting \cite{luo2020semi} to randomly select 62 scans for training, and 20 scans for evaluation. For pre-processing, we first rescale the intensities of the CT images into the window [$-125$, $275$] HU~\cite{zhou2019prior}, and then re-sample all the data to a isotropic resolution of $1.0\times1.0\times1.0 \text{mm}^3$. We crop all the scans at the pancreas region and normalized into zero and unit variance with the size of sub-volumes $96\times 96\times 96 \text{mm}^3$. In this study, we conduct all experiments on LA and pancreas dataset under 10\% and 20\% labeled ratios.

\begin{figure}[t]
	\centering
	\includegraphics[width=\linewidth]{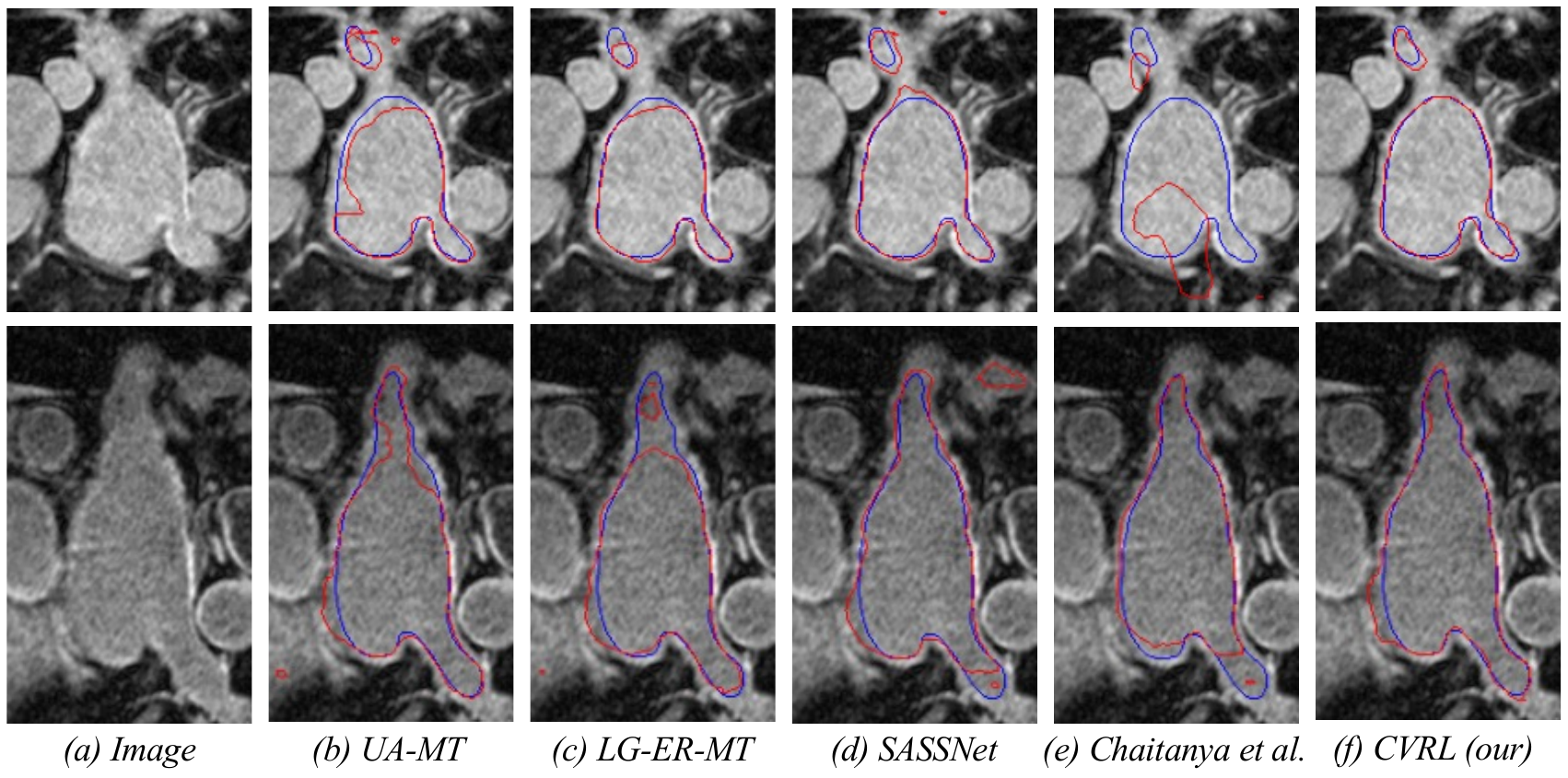}
    \vspace{-10pt}
	\caption{Visual comparisons with other methods. As observed, our CVRL achieves superior performance with more accurate borders and shapes.}
    \vspace{-10pt}
	\label{fig:vis2d}
\end{figure}

\myparagraph{Implementation Details}
In our framework, we use V-Net as the network backbone for two networks. For data augmentation, we use standard data augmentation techniques \cite{yu2019uncertainty,taleb20203d}. We empirically set the hyper-parameters $\alpha$,\,$\beta$,\,$\gamma$,\,$\tau$ as $0.1$,\,$0.1$,\,$0.1$,\,$1.0$, respectively. We use SGD optimizer with a momentum $0.9$ and weight decay $0.0005$ to optimize the network parameters. The initial learning rate is set as $0.01$ and divided by $10$ every $3000$ iterations. For EMA updates, we follow the experimental setting in \cite{yu2019uncertainty}, where the EMA decay rate $\alpha$ is set to $0.999$. We use the time-dependent Gaussian warming-up function~$\Psi_{con}(t)=\exp{\left(-5\left(1-t / t_{\max }\right)^{2}\right)}$ to ramp up parameters, where $t$ and $t_{max}$ denote the current and the maximum training step, respectively. For fairness, all evaluated methods are implemented in PyTorch, and trained for $10000$ iterations on an NVIDIA 3090Ti GPU with batch size $6$. In the testing stage, we adopt four metrics to evaluate the segmentation performance, including Dice coefficient (Dice), Jaccard Index (Jaccard), 95\% Hausdorff Distance (95HD), and Average Symmetric Surface Distance (ASD).

\myparagraph{Comparison with Other Semi-supervised Methods}
We evaluate our CVRL with several state-of-the-art semi-supervised segmentation methods on different amounts of labeled data, including V-Net \cite{milletari2016v}, DAN \cite{zhang2017deep}, DAP \cite{zheng2019semi}, UA-MT \cite{yu2019uncertainty}, LG-ER-MT \cite{hang2020local}, SASSNet \cite{li2020shape}, and Chaitanya \etal~\cite{chaitanya2020contrastive}. Table \ref{tab:la_16} compares our segmentation results with other methods.

We first conduct experiments under 20\% annotation ratios (16 labeled and 64 unlabeled). Under this setting, most above approaches achieve superior segmentation performance. CVRL gives better performance thanks to its low-level and high-level voxel-wise feature extraction. In particular, our proposed method outperforms other end-to-end semi-supervised methods in Dice (90.45\%), Jaccard (83.02\%), ASD (1.81), and 95HD (6.56).

To further evaluate the effectiveness of CVRL, we compare it with other methods in 10\% annotation ratio (8 labeled and 72 unlabeled), as reported in Table \ref{tab:la_16}. We observe consistent performance improvements over state-of-the-arts, in terms of Dice (88.56\%), and Jaccard (78.89\%). This evidence that i). taking voxel samples with contrastive learning yields better voxel embeddings; ii) both high-level and low-level relations are informative cues in both batch and feature dimension; iii) utilizing dimensional contrast is capable of consistently helping improve the segmentation performance. Leveraging all these aspects, it can observe consistent performance gains. As shown in Fig. \ref{fig:vis2d}, our method is capable of generating more accurate segmentation, considering the fact the improvement in such setting is difficult. This demonstrates i) the necessity of comprehensively considering both high-level and low-level contrast in both batch and feature dimension; and ii) efficacy of both inter-instance and intra-instance constraints. We also assess the performance of CVRL on Pancreas. We provide detailed evaluation results on Pancreas in Appendix Table \ref{table:pa}. We find that CVRL significantly outperforms all the state-of-the-art methods by a significant margin. We noticed that our proposed CVRL could improve results by an especially significant margin, with up to 3.25 - 5.21$\%$ relative improvement in Dice.

\begin{table}[t]
	\caption{Ablation study for the key component modules of CVRL on the LA dataset with 10\% annotation ratio (8 labeled and 72 unlabeled).}
	\label{tab:la_ablation}
	\resizebox{\textwidth}{!}{
		\begin{tabular}{l|c|c|c|c|c|c} \hline \hline
			\multirow{2}{*}{Method} & \multicolumn{2}{c|}{\# \textbf{scans used}} & \multicolumn{4}{c}{\textbf{Metrics}} \\ \cline{2-7}
			& Labeled        & Unlabeled        & Dice{[}\%{]} & Jaccard{[}\%{]} & ASD{[}voxel{]} & 95HD{[}voxel{]} \\ \hline
			Baseline           & 8 & 72  & 83.09 & 71.75 & 5.53 & 19.65 \\ \hline
			Baseline+$\mathcal{L}^{\text{high}}$ & 8 & 72 & 87.46 & 78.12 & 3.03 & 9.99 \\ \hline 
			Baseline+$\mathcal{L}^{\text{low}}$ & 8 & 72 & 87.24 & 77.49 & 3.36 & 10.13 \\ \hline
			Baseline+$\mathcal{L}^{\text{con}}$ & 8 & 72 & 85.72 & 75.31 & 4.72 & 13.18 \\ \hline 
			Baseline+$\mathcal{L}^{\text{high}}$+$\mathcal{L}^{\text{low}}$ & 8 & 72 & 88.14 & 78.38 & 3.02 & 9.58 \\ \hline
			Baseline+$\mathcal{L}^{\text{high}}$+$\mathcal{L}^{\text{low}}$+$\mathcal{L}^{\text{con}}$ & 8 & 72 & 88.56 & 78.89   &  2.81     & 8.22 \\ \hline
			\hline
		\end{tabular}
	}
	\vspace{-10pt}
\end{table}

\myparagraph{Ablation Study} 
We perform ablation experiments to validate the effectiveness of major components in our proposed method, including high-level and low-level contrastive strategy, and consistency loss. The quantitative results is reported in Table \ref{tab:la_ablation}. We compare CVRL with its five variants under 10\% annotation ratio (8 labeled and 72 unlabeled). Specially, the Baseline model refers to MT \cite{tarvainen2017mean}. We gradually incorporate $\mathcal{L}^{\text{high}}$, $\mathcal{L}^{\text{low}}$, $\mathcal{L}^{\text{con}}$, denoted as Baseline+$\mathcal{L}^{\text{low}}$, Baseline+$\mathcal{L}^{\text{high}}$, Baseline+$\mathcal{L}^{\text{con}}$, Baseline+$\mathcal{L}^{\text{low}}$+$\mathcal{L}^{\text{high}}$, Baseline+$\mathcal{L}^{\text{low}}$+$\mathcal{L}^{\text{high}}$+$\mathcal{L}^{\text{con}}$ (CVRL), respectively. As shown in the table, the Baseline network achieve 83.09\%, 71.75\%, 5.53, 19.65 in terms of Dice, Jaccard, ASD, and 95HD. With the progressive introduction of $\mathcal{L}^{\text{high}}$, $\mathcal{L}^{\text{low}}$, $\mathcal{L}^{\text{con}}$, our proposed algorithm enjoys consistently improvement gains over the Baseline network, boosting Dice and Jaccard by 5.47\%, 7.14\%, respectively. Also, the metrics ASD and 95HD are reduced by 2.72 and 11.43, respectively. This further validates the effectiveness of each key component. We summarize the effects of hyperparameters in Appendix Figure \ref{fig:vis_hyper}. 

\section{Conclusion}

In this work, we propose CVRL, a semi-supervised contrastive representation distillation framework by leveraging low-level and high-level cues to learn voxel-wise representations for volumetric medical image segmentation. Specifically, we propose to use voxel-wise contrastive and dimensional contrastive learning to ensure diversity promotion and exploit complex relations among training voxels. We also show the hardness-aware property is a key property for the success of our proposed dimensional contrastive learning. Experimental results demonstrate that our model yields state-of-the-art performance with generating more accurate boundaries with very limited annotations.

%
%
%
\bibliographystyle{splncs04}
\bibliography{ref}

\begin{thebibliography}{10}
\providecommand{\url}[1]{\texttt{#1}}
\providecommand{\urlprefix}{URL }
\providecommand{\doi}[1]{https://doi.org/#1}

\bibitem{bai2019self}
Bai, W., Chen, C., Tarroni, G., Duan, J., Guitton, F., Petersen, S.E., Guo, Y.,
  Matthews, P.M., Rueckert, D.: Self-supervised learning for cardiac mr image
  segmentation by anatomical position prediction. In: MICCAI. pp. 541--549.
  Springer (2019)

\bibitem{bortsova2019semi}
Bortsova, G., Dubost, F., Hogeweg, L., Katramados, I., de~Bruijne, M.:
  Semi-supervised medical image segmentation via learning consistency under
  transformations. In: MICCAI. pp. 810--818. Springer (2019)

\bibitem{chaitanya2020contrastive}
Chaitanya, K., Erdil, E., Karani, N., Konukoglu, E.: Contrastive learning of
  global and local features for medical image segmentation with limited
  annotations. In: NeurIPS (2020)

\bibitem{chen2019multi}
Chen, S., Bortsova, G., Ju{\'a}rez, A.G.U., van Tulder, G., de~Bruijne, M.:
  Multi-task attention-based semi-supervised learning for medical image
  segmentation. In: MICCAI. pp. 457--465. Springer (2019)

\bibitem{chen2020simple}
Chen, T., Kornblith, S., Norouzi, M., Hinton, G.: A simple framework for
  contrastive learning of visual representations. In: ICML. pp. 1597--1607.
  PMLR (2020)

\bibitem{hadsell2006dimensionality}
Hadsell, R., Chopra, S., LeCun, Y.: Dimensionality reduction by learning an
  invariant mapping. In: CVPR. vol.~2, pp. 1735--1742. IEEE (2006)

\bibitem{hang2020local}
Hang, W., Feng, W., Liang, S., Yu, L., Wang, Q., Choi, K.S., Qin, J.: Local and
  global structure-aware entropy regularized mean teacher model for 3d left
  atrium segmentation. In: MICCAI. pp. 562--571. Springer (2020)

\bibitem{hu2021semi}
Hu, X., Zeng, D., Xu, X., Shi, Y.: Semi-supervised contrastive learning for
  label-efficient medical image segmentation. In: MICCAI. Springer (2021)

\bibitem{hua2021feature}
Hua, T., Wang, W., Xue, Z., Ren, S., Wang, Y., Zhao, H.: On feature
  decorrelation in self-supervised learning. In: ICCV. pp. 9598--9608 (2021)

\bibitem{jing2021understanding}
Jing, L., Vincent, P., LeCun, Y., Tian, Y.: Understanding dimensional collapse
  in contrastive self-supervised learning. arXiv preprint arXiv:2110.09348
  (2021)

\bibitem{laine2016temporal}
Laine, S., Aila, T.: Temporal ensembling for semi-supervised learning. arXiv
  preprint arXiv:1610.02242  (2016)

\bibitem{li2020shape}
Li, S., Zhang, C., He, X.: Shape-aware semi-supervised 3d semantic segmentation
  for medical images. In: MICCAI. pp. 552--561. Springer (2020)

\bibitem{li2018semi}
Li, X., Yu, L., Chen, H., Fu, C.W., Heng, P.A.: Semi-supervised skin lesion
  segmentation via transformation consistent self-ensembling model. arXiv
  preprint arXiv:1808.03887  (2018)

\bibitem{luo2020semi}
Luo, X., Chen, J., Song, T., Wang, G.: Semi-supervised medical image
  segmentation through dual-task consistency. In: AAAI (2020)

\bibitem{milletari2016v}
Milletari, F., Navab, N., Ahmadi, S.A.: V-net: Fully convolutional neural
  networks for volumetric medical image segmentation. In: 3DV. pp. 565--571.
  IEEE (2016)

\bibitem{misra2020self}
Misra, I., Maaten, L.v.d.: Self-supervised learning of pretext-invariant
  representations. In: CVPR. pp. 6707--6717 (2020)

\bibitem{nie2018asdnet}
Nie, D., Gao, Y., Wang, L., Shen, D.: Asdnet: Attention based semi-supervised
  deep networks for medical image segmentation. In: MICCAI. pp. 370--378.
  Springer (2018)

\bibitem{oord2018representation}
Oord, A.v.d., Li, Y., Vinyals, O.: Representation learning with contrastive
  predictive coding. arXiv preprint arXiv:1807.03748  (2018)

\bibitem{roth2016data}
Roth, H.R., Farag, A., Turkbey, E., Lu, L., Liu, J., Summers, R.M.: Data from
  pancreas-ct. the cancer imaging archive (2016)

\bibitem{taleb20203d}
Taleb, A., Loetzsch, W., Danz, N., Severin, J., Gaertner, T., Bergner, B.,
  Lippert, C.: 3d self-supervised methods for medical imaging. In: NeurIPS. pp.
  18158--18172 (2020)

\bibitem{tarvainen2017mean}
Tarvainen, A., Valpola, H.: Mean teachers are better role models:
  Weight-averaged consistency targets improve semi-supervised deep learning
  results. In: NeurIPS. pp. 1195--1204 (2017)

\bibitem{tian2019contrastive}
Tian, Y., Krishnan, D., Isola, P.: Contrastive multiview coding. arXiv preprint
  arXiv:1906.05849  (2019)

\bibitem{wang2021understanding}
Wang, F., Liu, H.: Understanding the behaviour of contrastive loss. In: CVPR.
  pp. 2495--2504 (2021)

\bibitem{wang2015unsupervised}
Wang, X., Gupta, A.: Unsupervised learning of visual representations using
  videos. In: ICCV. pp. 2794--2802 (2015)

\bibitem{wu2018unsupervised}
Wu, Z., Xiong, Y., Stella, X.Y., Lin, D.: Unsupervised feature learning via
  non-parametric instance discrimination. In: CVPR (2018)

\bibitem{you2021simcvd}
You, C., Zhou, Y., Zhao, R., Staib, L., Duncan, J.S.: Simcvd: Simple
  contrastive voxel-wise representation distillation for semi-supervised
  medical image segmentation. arXiv preprint arXiv:2108.06227  (2021)

\bibitem{yu2019uncertainty}
Yu, L., Wang, S., Li, X., Fu, C.W., Heng, P.A.: Uncertainty-aware
  self-ensembling model for semi-supervised 3d left atrium segmentation. In:
  MICCAI. pp. 605--613. Springer (2019)

\bibitem{zhang2017deep}
Zhang, Y., Yang, L., Chen, J., Fredericksen, M., Hughes, D.P., Chen, D.Z.: Deep
  adversarial networks for biomedical image segmentation utilizing unannotated
  images. In: MICCAI. pp. 408--416. Springer (2017)

\bibitem{zheng2019semi}
Zheng, H., Lin, L., Hu, H., Zhang, Q., Chen, Q., Iwamoto, Y., Han, X., Chen,
  Y.W., Tong, R., Wu, J.: Semi-supervised segmentation of liver using
  adversarial learning with deep atlas prior. In: MICCAI. pp. 148--156.
  Springer (2019)

\bibitem{zhou2019prior}
Zhou, Y., Li, Z., Bai, S., Wang, C., Chen, X., Han, M., Fishman, E., Yuille,
  A.L.: Prior-aware neural network for partially-supervised multi-organ
  segmentation. In: ICCV. pp. 10672--10681 (2019)

\end{thebibliography}

\end{document}